\documentclass[runningheads]{llncs}
\pdfoutput=1
\usepackage{eccv}

\usepackage{eccvabbrv}

\usepackage{graphicx}
\usepackage{booktabs}
\usepackage{multirow}
\usepackage{marvosym}

\usepackage[accsupp]{axessibility}  % Improves PDF readability for those with disabilities.

\usepackage[pagebackref,breaklinks,colorlinks,citecolor=eccvblue]{hyperref}
\usepackage{orcidlink}

\begin{document}

\title{Towards Full-parameter and Parameter-efficient Self-learning For Endoscopic Camera Depth Estimation} 

\titlerunning{Endoscopic Camera Depth Estimation}

\author{Shuting Zhao\inst{1\star} \and
Chenkang Du\inst{2\star} \and
Kristin Qi\inst{3} \and
Xinrong Chen\inst{1}\textsuperscript{\Letter}\and
Xinhan Di\inst{2}\textsuperscript{\Letter}
}

\authorrunning{Zhao, Du et al.}
% Shanghai Key Laboratory of Medical Image Computing and Computer Assisted Intervention, Fudan University
\institute{Academy for Engineering \& Technology, Fudan University, Shanghai, China \email{\{zhaoshuting,chenxinrong\}@fudan.edu.cn}\\
\and
Giant Network AI Lab
\email{\{duchenkang,dixinhan\}@ztgame.com}\\
\and
University of Massachusetts, Boston, USA \email{yanankristin.qi001@umb.edu}\\
}

\maketitle
\begin{abstract}
Adaptation methods are developed to adapt depth foundation models to endoscopic depth estimation recently. However, such approaches typically under-perform training since they limit the parameter search to a low-rank subspace and alter the training dynamics. Therefore, we propose a full-parameter and parameter-efficient learning framework for endoscopic depth estimation. At the first stage, the subspace of attention, convolution and multi-layer perception are adapted simultaneously within different sub-spaces. At the second stage, a memory-efficient optimization is proposed for subspace composition and the performance is further improved in the united sub-space. Initial experiments on the SCARED \cite{allan2021stereo} dataset demonstrate that results at the first stage improves the performance from $10.2\%$ to $4.1\%$ for Sq Rel, Abs Rel, RMSE and RMSE log \cite{cui2024surgical,shao2022self,yang2024depth,yang2024self} in the comparison with the state-of-the-art models.
  \keywords{Depth Foundation Model \and Endoscopic Depth Estimation \and Full-Parameter and Efficient Learning}
\end{abstract}
\section{Introduction}
\label{sec:intro} 
Recently, attention is attracted on the foundation models for their good performances in a variety of tasks including text and vision \cite{huang2024endo,kirillov2023segment,oquab2023dinov2}. Then, the adaption of foundation models to the medical domain is developed for the image segmentation, detection and depth estimation \cite{chen2023sam,wu2023self,zhang2023customized}. However, such approaches typically under-perform training. Therefore, we propose a full-parameter and memory-efficient module connecting different sub-spaces to a united space for the adaption of the depth foundation model. 
\begin{figure}
\centering
\includegraphics[width=1.0\textwidth]{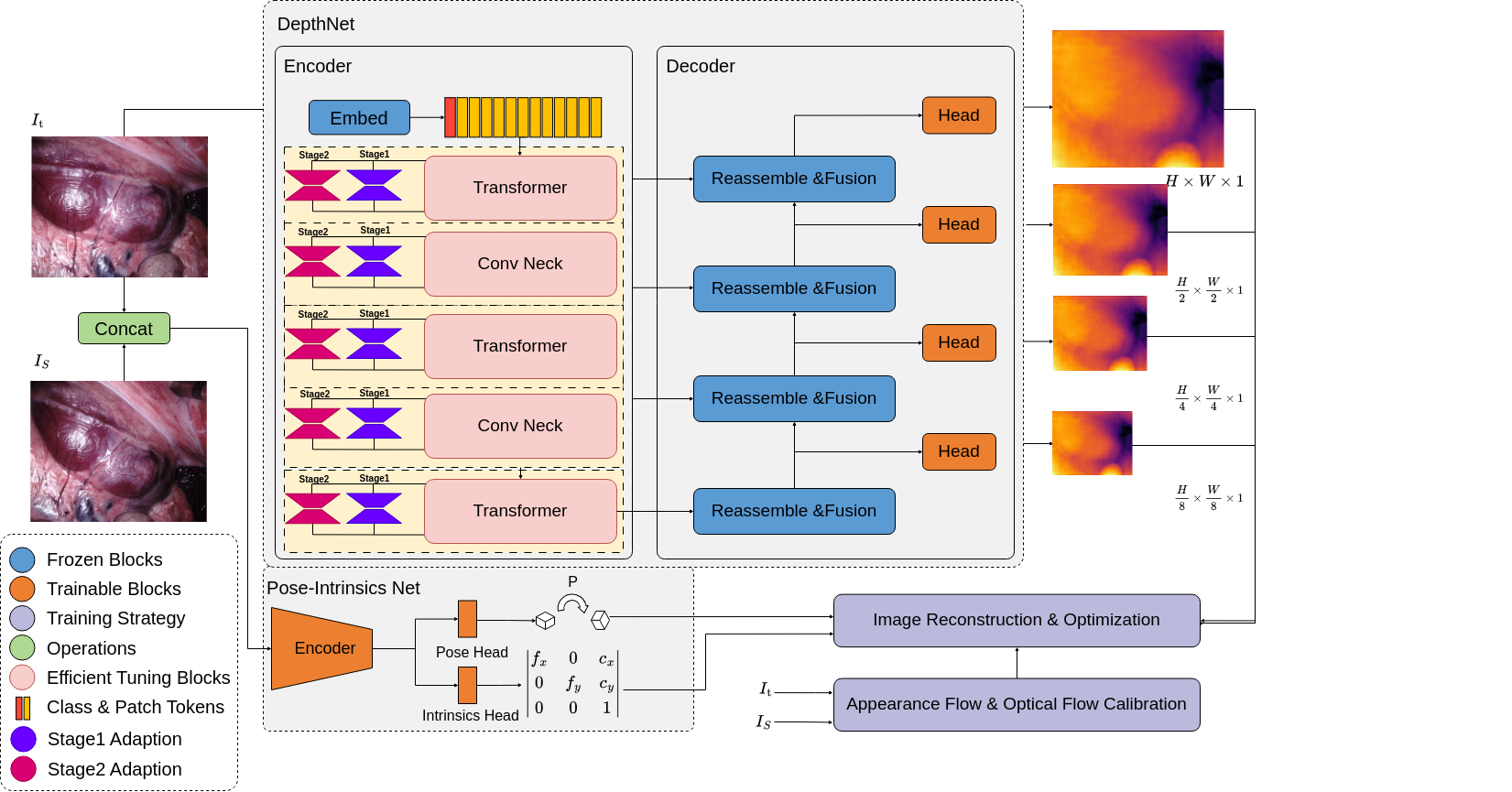}
\caption{Two-Stage Adaption on the Depth Foundation Model \cite{cui2024endodac}.}
\label{fig:1}
\end{figure}
\section{Related work}
Foundation Models are generally trained on extensive amounts and demonstrate strong generalization capabilities across multiple tasks and scenarios. For example, Depth Anything (DA) \cite{yang2024depth} is a depth estimation foundation model trained on large-scale labeled and unlabeled data. However, the adaption should be conducted on these foundation models for the endoscopic scenes. Then, the adaption of foundation models to medical domain is developed such as segmentation, detection and depth estimation \cite{zhang2023customized,cui2024endodac}. The majority of these approaches are in the field of low-rank adaption \cite{hu2021lora}. However, this adaption is limited in the single sub-space. Therefore, we proposed a full-parameter and memory-efficient module connecting different sub-spaces and project to a united space.
\begin{figure}
\centering
\includegraphics[width=1.0\textwidth]{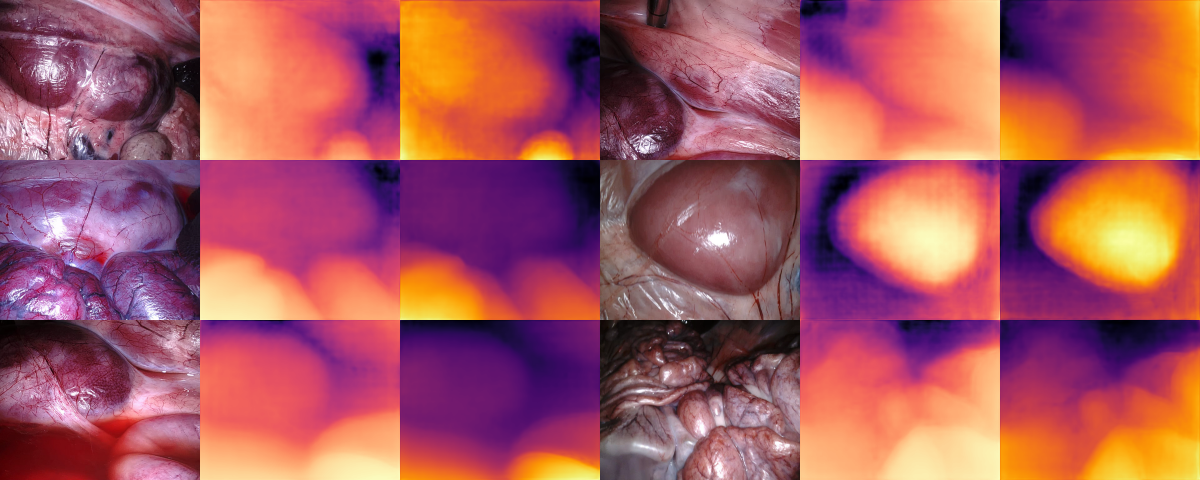}
\caption{The first and fourth column represent GT RGB images. The second and the fifth column represent the depth visualization of the state-of-the-art model \cite{cui2024endodac}. The third and the sixth column represent the depth visualization of the proposed first stage module.}
\label{fig:2}
\end{figure}
\section{Methods}
\label{sec:method}
We propose a two-stage adaption strategy (Figure \ref{fig:1}) for the adaption of the state-of-the-art depth foundation model \cite{cui2024endodac}. At the first stage, a multiple number of adapters are applied to different sub-spaces of the foundation model. At the second stage, a bridge is built to combine different sub-spaces into a united space and the performance is continued to be improved with efficient memory. In details, we represent the state-of-the-art depth model \cite{cui2024endodac} as three types of sub-spaces, the convolution space, the mlp space and the attention space. It's represented as the following:
\begin{equation}
W_{depth} = W_{conv} \cup W_{mlp} \cup W_{atten}
\end{equation}
where $W_{conv} = W_{conv}^{1} \cup W_{conv}^{2} \cup \dots W_{conv}^{n_{1}}$, representing weights of $n_1$ number of convolution layers, $W_{mlp} = W_{mlp}^{1} \cup W_{mlp}^{2} \cup \dots W_{mlp}^{n_{2}}$, representing weights of $n_2$ number of mlp layers, $W_{atten} = W_{atten}^{1} \cup W_{atten}^{2} \cup \dots W_{atten}^{n_{3}}$, representing weights of $n_3$ number of attention layers.
At the first stage, low-rank updates \cite{cui2024endodac} are developed for each of the above weight as the following:
\begin{equation}
W_{i}^{stage1} = W_{i} + B_{i}A_{i}
\end{equation}
where $W_{i}$ represents the weights of each layer in $W_{depth}$, $i \in \{1,2,\dots n_{1} + n_{2} + n_{3}\}$ $W_{i} \in \mathcal{R}^{m \times n}$, $B_{i} \in \mathcal{R}^{m \times n}$ and $A_{i} \in \mathcal{R}^{r \times n}$, and $r \ll min(m,n)$. $A_{i}$ and $B_{i}$ are the learnable low-rank adapters and $W_{i}$ is a fixed weight matrix.
Then, at the second stage, a bridge is built through the projection of gradient to combine different sub-spaces into a unified space with efficient memory \cite{cui2024endodac}. It's represented as the following: 
\begin{equation}
B_{i}^{stage2}  = - \Delta_{W^{i}} (W^{i}) 
\end{equation}
Therefore, the full-parameter adaption for each layer is represented as the following:
\begin{equation}
W_{i}^{stage2} = \alpha \times W_{i}^{stage1} + \beta \times B_{i}^{stage1}
\end{equation}
For each type of sub-spaces, the module is consisted of the low-rank weight adaption part and the full-parameter gradient adaption part. To be noted, $\alpha$ and $\beta$ are learnable parameters.
\begin{table}
\begin{center}
\begin{tabular}{|c|c|c|c|c|c|c|}
 \hline
 \multicolumn{7}{|c|}{SCARED Dataset \cite{allan2021stereo}.} \\
 \hline
 Method &Abs Rel &Sq Rel &RMSE &RMSE log & $\delta$ & Total(M)\\
 \hline
 Fang \cite{fang2020towards}  &0.078  &0.794 &6.794 &0.109 &0.946 & 136.8\\
 \hline
 %Defeat-Net \cite{spencer2020defeat} &0.077  &0.792 &6.688 &0.108 &0.941 & 14.8\\
 %\hline
 %SC-SFMLearner \cite{bian2019unsupervised} &0.068  &0.645 &5.988 &0.097 &0.957 & 14.8\\
 %\hline
 Monodepth2 \cite{godard2019digging} &0.069  &0.577 &5.546 &0.094 &0.948 & 14.8\\
 \hline
 Endo-SfM \cite{ozyoruk2021endoslam} &0.062  &0.606 &5.726 &0.093 &0.957 & 14.8\\
 \hline
 AF-SFMLearner \cite{shao2022self} &0.059  &0.435 &4.925 &0.082 &0.974 & 14.8\\
 \hline
 Yang \cite{yang2024self} &0.062  &0.558 &5.585 &0.090 &0.962 & 2.0\\
 \hline
 DA \cite{yang2024depth} &0.058 &0.451 &5.058 &0.081 &0.974 & 97.5\\
 \hline
 EndoDAC \cite{cui2024endodac} &0.052  &0.362 &4.464 &0.073 &0.979 & 99.0\\
 \hline
 Ours(First-Stage) &\textbf{0.049}  &\textbf{0.325} & \textbf{4.280} & \textbf{0.069} & \textbf{0.983} & 99.1 \\
 \hline
\end{tabular}
\end{center}
\caption{Quantitative depth comparison on SCARED \cite{allan2021stereo} dataset of SOTA self-supervised learning depth estimation methods. The best results are in bold.}
\label{tab:1}
\end{table}
\section{Experiments}
\label{sec:exp}
\subsubsection{SCARED Dataset \cite{allan2021stereo}.}
SCARED \cite{allan2021stereo} contains $35$ endoscopic videos with $22950$ frames of fresh porcine cadaver abdominal anatomy collected with a da Vinci Xi endoscope. We followed the split scheme where the SCARED dataset \cite{allan2021stereo} is split into $15351$, $1705$, and $551$ frames for the training, validation and test sets, respectively.
\subsubsection{Evaluation Settings.}
Following \cite{cui2024surgical,shao2022self,yang2024depth,yang2024self}, we compute the $5$ standard metrics: Abs Rel, Sq Rel, RMSE, RMSE log and $\delta$ for evaluation. We re-scale the predicted depth map by a median scaling method \cite{cui2024surgical,shao2022self,zhou2017unsupervised} during evaluation. The first stage of the adaption module is evaluated in our initial experiment. It is in the comparison with the state-of-the-art of the depth estimation. models\cite{recasens2021endo,shao2022self,yang2024self}. The result (Table \ref{tab:1}) demonstrates that it reduces the Abs Rel by $5.7\%$(from $0.052$ to $0.049$), Sq Rel by $10.2\%$(from $0.362$ to $0.325$), RMSE by $4.1\%$(from $4.464$ to $4.280$), RMSE log by $5.8\%$(from $0.073$ to $0.069$), rises the $\delta$ by $0.4\%$(from $0.979$ to $0.983$). Besides, ablation studies are conducted on the different modules. As presented in Table \ref{tab:2}, the ablation studies demonstrate the effectiveness of each sub-space. The qualitative depth estimation is in comparison with the state-of-the-art model(Figure \ref{fig:2}). From the visualization, the proposed adaption generates a more accurate geometry relation within the depth map.
\begin{table}
\begin{center}
\begin{tabular}{|c|c|c|c|c|c|}
 \hline
 \multicolumn{6}{|c|}{Ablation Study of Adaption on the First Stage.} \\
 \hline
 Method &Abs Rel &Sq Rel &RMSE &RMSE log & $\delta$ \\
 \hline
 MLP-Space   &0.051  &0.362 &4.552 &0.073 &0.982 \\
 \hline
 MLPA+ConvA-Space &0.050  &0.332 &4.346 &0.071 &0.982 \\
 \hline
 MLPA+ConvA+AttnA-Space &\textbf{0.049}  &\textbf{0.325} & \textbf{4.280} & \textbf{0.069} & \textbf{0.983} \\
 \hline
\end{tabular}
\end{center}
\caption{Ablation Study on SCARED \cite{allan2021stereo} dataset.}
\label{tab:2}
\end{table}   
\section{Discussion}
\label{sec:discussion}
We propose a two-stage adaption for the depth foundation model towards full-parameter with efficient memory. Experiments of the first stage are conducted and the results demonstrate that the error reduction is from $10.2\%$ to $4.1\%$ for Sq Rel, Abs Rel, RMSE and RMSE log \cite{cui2024surgical,shao2022self,yang2024self}. Then, the experiments for the second stage are planed to build a bridge to unify different types of the sub-space with efficient memory. Finally, we are exploring the third stage to combine different depth foundation model with efficient parameters for improving the performance further. 

\bibliographystyle{splncs04}
\bibliography{main}
\end{document}